\documentclass[a4paper]{article}
\usepackage{algorithm}
\usepackage[noend]{algpseudocode}
\usepackage[utf8]{inputenc}
\usepackage{INTERSPEECH2019}
\usepackage{float}
\usepackage{amsmath}
\usepackage{amssymb}
\usepackage{array}

\usepackage{todonotes}
\newcolumntype{P}[1]{>{\centering\let\newline\\\arraybackslash\hspace{0pt}}m{#1}}

\title{M2H-GAN: A GAN-based Mapping from Machine to Human Transcripts \\ for Speech Understanding}

\name{Titouan Parcollet$^{1,2}$, Mohamed Morchid$^1$, Xavier Bost$^2$, Georges Linarès$^1$}
\address{
  $^1$Avignon Université, LIA, France \\
  $^2$ORKIS, Aix-en-provence, France }
\email{titouan.parcollet@alumni.univ-avignon.fr, \{firstname.lastname\}@univ-avignon.fr}

\begin{document}

\maketitle
\begin{abstract}
  
  Deep learning is at the core of recent spoken language understanding (SLU) related tasks. More precisely, deep neural networks (DNNs) drastically increased the performances of SLU systems, and numerous architectures have been proposed. In the real-life context of theme identification of telephone conversations, it is common to hold both a human, manual (TRS) and an automatically transcribed (ASR) versions of the conversations. Nonetheless, and due to production constraints, only the ASR transcripts are considered to build automatic classifiers. TRS transcripts are only used to measure the performances of ASR systems. Moreover, the recent performances in term of classification accuracy, obtained by DNN related systems are close to the performances reached by humans, and it becomes difficult to further increase the performances by only considering the ASR transcripts. This paper proposes to distillates the TRS knowledge available during the training phase within the ASR representation, by using a new generative adversarial network called M2H-GAN to generate a TRS-like version of an ASR document, to improve the theme identification performances.

\end{abstract}
\noindent\textbf{Index Terms}: Spoken language understanding, generative adversarial networks.

\section{Introduction}

Spoken language understanding (SLU) has been massively impacted by machine learning (ML) algorithms, and more precisely by deep neural networks (DNNs). Interesting solutions have been therefore proposed for SLU in human-computer and human-human dialogues \cite{sarikaya2014application,serban2016building,chen2016end,serdyuk2018towards}. An important component of this domain area is the task of topic identification in text documents \cite{tur2011spoken}. As an example, this paper deals with customer care services (CCS) in which an agent interact with a customer to address her/his concerns and to provide a solution. The automatic system is expected to correctly identifies the major theme of the conversation from the transcriptions obtained by an automatic speech recognition system (ASR), or by human manual transcripts (TRS). Unfortunately, TRS versions are only available at training time due to the fact that the production environment is automated, and relies on ASR transcripts. To address this problem, various neural architectures have been developed to directly classify the ASR transcriptions of telephone conversations, based on multi-layer perceptrons and pre-trained deep neural networks \cite{parcollet2016quaternion,parcollet2017deep}, or convolutional neural networks \cite{parcollet2018quaternion}. Nonetheless, while such models are powerful, they are also limited by the quality of the ASR transcriptions. \cite{janod2017denoised,Janod+2016} proposed to use the knowledge available at training time through the TRS transcriptions to enhance the input representation of the ASR versions. This enhancement is made possible with the use of stacked and deep stacked auto-encoders to learn a static mapping that projects the ASR latent space to the TRS one. Based on the promising results observed with this approach, we propose to further investigate the distillation of the TRS knowledge to the ASR representation with the recent generative adversarial networks (GAN). 

GANs are an active field of research and offer an interesting approach that focuses on a game-theoretic method to train a generative model \cite{goodfellow2014generative}. Numerous architectures have been proposed to address various tasks \cite{radford2015unsupervised,wu2017adversarial}. From a simplified perspective, GANs are commonly used to learn a mapping from a random noise space to a target one, making it possible to generate new unseen samples. In natural language processing (NLP) tasks, the noise space is commonly replaced with a well defined input representation, such as text written in a specific language for neural machine translation \cite{yang2017improving}. Then, GANs are used to project this latent representation to a different target language. In the task of theme identification of telephone conversations investigated in this paper, we consider the latent ASR transcription as the noise space, and the TRS versions as the target one. After training, the model is expected to enhance the ASR latent representation with TRS knowledge, to further improve the results when classifying the documents.   

This paper proposes a task adapted model called Machine-to-Human GAN (M2H-GAN) by merging the GAN with a semi-supervised GAN (SGAN), to better represent and classify telephone conversations. Therefore, the contributions of the paper are:
\begin{itemize}
    \item Introduce a new GAN architecture called M2H-GAN to efficiently map the automatically transcribed representation of conversations, to a latent representation of their manually transcribed version (Section \ref{sec:generative}).
    \item Compare the classification accuracy obtained with this new representation to other methods on a theme identification of telephone conversations task (Section \ref{sec:exps}).
\end{itemize}
The experiments conduced on the DECODA \cite{bechet2012decoda} dataset show that the M2H-GAN reduces the performances gap in term of classification accuracy between automatically and manually transcribed documents by learning a robust mapping between the two latent sub-spaces. The M2H-GAN also offers a more stable classification process with a lowered standard deviation with respect to results observed. 

\begin{figure*}[!t]
    \centering
    \includegraphics[scale=0.065]{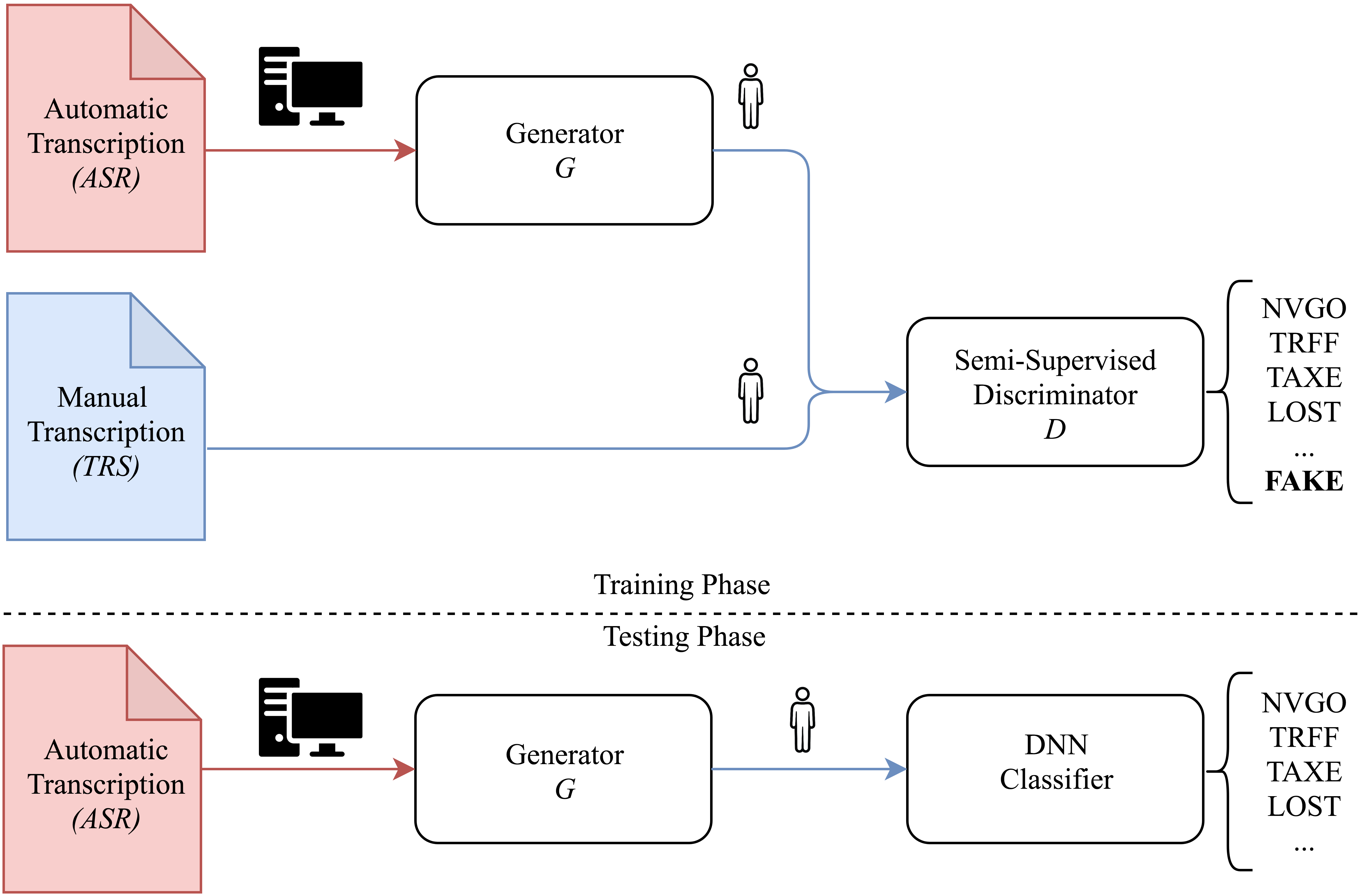}
    \caption{Illustration of the M2H-GAN architecture at training (top) and testing (bottom) time. Red and blue lines show the ASR and TRS representation signal. Note that the output of the generator $G$ goes from red to blue during the training phase. }
    \label{fig:M2H}
\end{figure*}

\section{Related work}

Generative neural models are a specific and active domain area in the machine learning field. Recently, generative adversarial networks (GANs) \cite{goodfellow2014generative} received an astonishing interest due to the remarkable results obtained in computer vision \cite{zhu2017unpaired,radford2015unsupervised,kim2017learning,berthelot2017began}. The ability of GANs to generate samples that are closely-related to targets ones has also been extended to natural language processing (NLP) tasks, such as text and dialogue generation \cite{rajeswar2017adversarial,donahue2018adversarial,li2017adversarial}, or neural machine translation \cite{yang2017improving,wu2017adversarial}. To the best of our knowledge, \cite{wu2017adversarial} is the most related work to the problem addressed in this paper. Indeed, \cite{wu2017adversarial} proposed a model aiming to generate sentences which are hard to be discriminated from human-translated sentences. Consequently, the GAN model is expected to learn a mapping from one language to another one based on human manual translations. The task considered in this paper replaces the initial language by an automatic transcription of a conversation, and the target language by its manual transcription. Furthermore, we propose to perform classification on top of the generation, as driven by the semi-supervised GAN (SGAN) approach \cite{odena2016semi}. Nonetheless, our model uses a different architecture that does not take into consideration the target classes when generating the samples, since \textit{golden-targets} (i.e manual transcriptions) are not available at testing time.

\section{Generative neural models}
\label{sec:generative}

In this paper a basic GAN is merged with the semi-supervised SGAN (Section \ref{subsec:sgan}) to allow a projection of an automatically transcribed document, to its manual transcription representation with the Machine-to-Human GAN (M2H-GAN, Section \ref{subsec:M2H}).  

\subsection{Generative Adversarial Networks}
\label{subsec:sgan}

In a generative adversarial network \cite{goodfellow2014generative}, two neural networks are trained in opposition. First, a generator $G$ outputs a fake object named $\Tilde{x}$ from an input random noise vector $z$:

\begin{equation}
    \Tilde{x} = G(z)
\end{equation}

Then, a discriminator $D$ receives alternatively a true sample $x$ or a fake one $\Tilde{x}$ from $G$, and outputs a probability distribution of the input being a fake or not. During training, $D$ tries to maximize the log-likelihood of the correctly assigned source:

\begin{equation}
    L = \mathbb{E}[\log p(\text{real} | x)] + \mathbb{E}[\log p(\text{fake} | \Tilde{x})]
\label{eq:loss}
\end{equation}

In the same manner, $G$ is trained to fool $D$ by minimizing the second term of Eq.~\ref{eq:loss}. Indeed, reducing the probability of correct classification of fake inputs increases the generating capability of $G$. 

Auxiliary and semi-supervised GANs \cite{mirza2014conditional,odena2016semi} have been proposed to take into consideration the labels in both the generator and the discriminator to drive the generation process toward a specific class. In an SGAN, $D$ is trained to determine if the input signal is fake or of a certain label. Consequently, the output dimension of $D$ is of size $N+1$ with $N$ being the number of classes, and $+1$ representing the \textit{fake} case. The loss function remains unchanged. SGANs use labels to add a condition on the generation process, making it possible to generate samples of a specific class, such as \textit{car} or \textit{bird} for image generation.  

\subsection{Machine-to-Human representation with generative models}
\label{subsec:M2H}

We propose to merge the initial GAN with its semi-supervised version SGAN, in a model named Machine-to-Human GAN(M2H-GAN). An overview of the M2H-GAN architecture is depicted in Figure \ref{fig:M2H}. In M2H-GAN, $\Tilde{x}$ is the generated representation of an automatically transcribed document (ASR) from $G$, and $x$ is the ``clean'' TRS version of the same sample. $D$ is trained to determine if the input has been generated, or belongs to a certain class (SGAN), and thus contains $N+1$ output neurons. Consequently, $G$ is jointly trained to map the ASR representation to a latent TRS and ``clean'' representation, in order to fool the discriminator. Unlike for SGAN, the generator of M2H-GAN does not have access to the label, due to the fact that conversations classes are unknown during the testing phase. This modification allows the discriminator to have more room to discover if an input is fake or not, making it more powerful. As a consequence, the generator must create a more convincing representation of the ASR signal, and receives gradients according to the label, without any conditioning on the input. An overview of M2H-GAN is depicted in Figure \ref{fig:M2H}.

\section{Experiments}
\label{sec:exps}

This section introduces the theme classification of telephone conversations task with the DECODA dataset (Section \ref{subsect:decoda}), alongside with the proposed representation of the document (Section \ref{subsec:lda}). The investigated architectures are detailled in Section \ref{subsec:modelsarchitectures}, while the observed results are reported in Section \ref{subsec:results}.

\subsection{Spoken conversations dataset}
\label{subsect:decoda}

\begin{figure}[H]
\begin{center}
\includegraphics[scale=0.30]{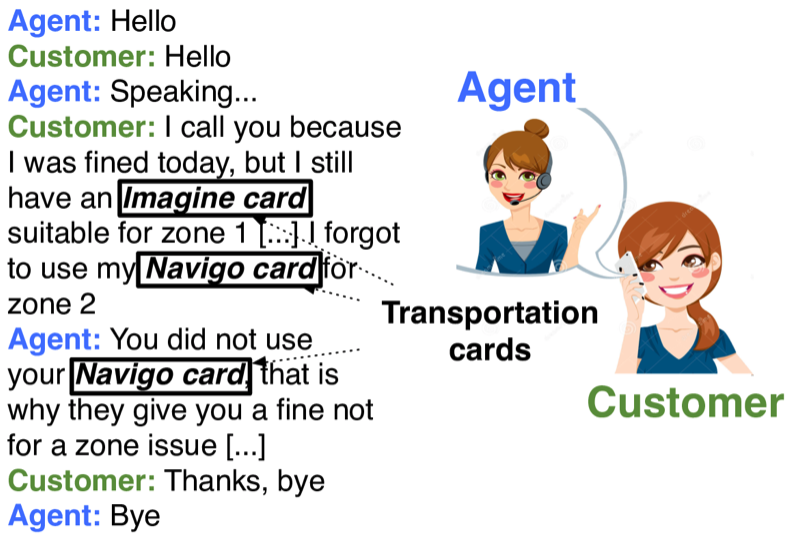}
 \caption{Example of a human transcription of a dialogue from the DECODA corpus for the SLU task of theme identification.}
 \label{fig:decoda}
\end{center}
\end{figure}

The corpus of spoken conversations is a set of automatically transcribed and annotated human-human telephone conversations of the Paris transportation system CCS (RATP). This corpus comes from the first version of the DECODA project~\cite{bechet2012decoda} and is employed to evaluate the effectiveness of the proposed M2H-GAN on a conversation theme identification task. The DECODA corpus is composed of $1,242$ telephone conversations recorded during high traffics days in the capital, which is equivalent to about $74$ hours of signal. The dataset was split into $8$ categories or dominant themes that are detailed in Table \ref{table:Decoda_Dataset}. An example of a manually transcribed conversation of DECODA is given in Figure \ref{fig:decoda}.

It is important to highlight the difficulty of the classification task due to the close sub-topics that can occur within a conversation. Indeed, a customer can ask for the price of a transportation card after a loss, and the document will be assigned to \textit{transportation cards}, while the vocabulary is also closely related to \textit{lost and found}. Furthermore, high word error rates (WER) are reported on the ASR transcripts with the LIA-Speeral ASR system \cite{linares2007lia}, due to very difficult and noisy environments including streets, buses and metros. Indeed, WERs of $45.8$\%, $59.3$\% and $58.0$\% are obtained on the training, validation and test sets respectively. Considering the high WERs and the closely related sub-topics within a document, it is crucial to introduce the clean and manual transcription of the conversation information to the training process, to build better classification systems.

\begin{table}[!h]
\centering
\centering
\caption{\label{table:Decoda_Dataset}DECODA dataset.}
\scalebox{0.95}{
\begin{tabular}{c c c c}
\hline
{\bf Class }&\multicolumn{3}{c}{{\it Number of samples}} \\
\cline{2-4}
{\bf label} & {\bf training} & {\bf development}  & {\bf testing}\\
\hline
problems of itinerary & 145 & 44 & 67 \\
lost and found & 143 & 33 & 63  \\
time schedules & 47 & 7 & 18  \\
transportation cards & 106 & 24 & 47  \\
state of the traffic & 202 & 45 & 90  \\
fares & 19 & 9 & 11  \\
infractions & 47 & 4 & 18 \\ 
special offers & 31 & 9 & 13  \\
\hline
\hline
{\bf Total} & {\bf 740} & {\bf 175} & {\bf327} \\
\hline
\end{tabular}
}
\end{table}

\subsection{Abstract document representation with LDA}
\label{subsec:lda}

The latent Dirichlet allocation or LDA is an effective method to represent documents in an unsupervised manner, as probability distributions of hidden topics \cite{blei2003latent} in a document, and have shown their efficiency in many previous related works \cite{morchid2013quaternions,parcollet2016quaternion}. For the experiments described in this section, LDA models are trained over the training set of DECODA following the standard hyper-parameters heuristic \cite{blei2003latent}. It is important to note that two LDA models are trained with either the ASR or TRS conversations from the training sub-set of the DECODA data-set. Consequently, $\alpha=\frac{50}{T}$, with $T$ the number of topics, and $\beta = 0.01$. The number $T$ has been previously investigated for this task in \cite{parcollet2016quaternion,parcollet2017deep}, and is set to $25$. More precisely, $10$ runs of the $T=25$ LDA model are concatenated to obtain a final vector of size $25\times10=250$, to alleviate any variations.  Then, every conversation is projected into the corresponding LDA space, and is embedded in a vector of size $250$. 

\subsection{Experimental protocol}
\label{subsec:modelsarchitectures}

To evaluate the effectiveness of the M2H-GAN to generate TRS-like representations of ASR transcripts, we compare M2H-GAN to a GAN model on the theme classification of telephone conversations. Deep feed-forward NNs trained on TRS and ASR transcripts are used as baselines. We also compare M2H-GAN to previously investigated generative models \cite{Janod+2016}. Training and testing steps are detailed in Algorithm 1, and can be summarized as follows: 1) Train GAN or M2H-GAN models; 2) Freeze the generator and train a DNN classifier from the generated features. Finally, Figure \ref{fig:M2H} represents the global architecture of the model.  

\begin{algorithm}[!h]
\label{algo:train}
    \caption{Training procedures.}
    \label{euclid}
    \begin{algorithmic}[1] 
        \Procedure{Train GANs}{$X_{trs}$,$X_{asr}$} 
        \State Project $X_{trs}$, $X_{asr}$ in LDA to obtain $Z_{trs}$, $Z_{asr}$.
        \State Generate $\Tilde{X}_{asr}$ with $G$ from $Z_{asr}$.
        \State Train $D$ and $G$ based on $\Tilde{X}_{asr}$, $Z_{trs}$. \cite{goodfellow2014generative}. 
        \EndProcedure
        \Procedure{Train DNNs}{$X_{asr}$} 
        \State Project $X_{asr}$ in LDA to obtain $Z_{asr}$.
        \State Generate $\Tilde{X}_{asr}$ with frozen $G$ from $Z_{asr}$.
        \State Train a DNN to classify $\Tilde{X}_{asr}$.
        \EndProcedure
    \end{algorithmic}
\end{algorithm}

\noindent\textbf{DNNs.} Classifiers rely on $2$ hidden layers of size $256$ with $\tanh{}$ activations, and a final \textit{softmax} layer corresponding to the $8$ themes of the DECODA dataset \cite{bechet2012decoda}. They are trained during $40$ epochs based on the Adam optimizer \cite{kingma2014adam} with vanilla hyper-parameters and no regularization techniques. After training, the maximum accuracy obtained on the test, alongside with the best result \textit{w.r.t} to the best validation performances are saved. 

\noindent\textbf{GAN.} The generator is made of $2$ hidden layers of size $512$ and $250$ (corresponding to the size of the LDA vector) with layer-wise normalization \cite{ba2016layer} and $\tanh{}$ activations, while the discriminator is composed of $2$ layers of $128$ and $8$ neurons with $\tanh{}$ and \textit{sigmoid} activation functions. The discriminating labels are smoothed by being sampled from a uniform distribution bounded by $[0.0,0.7]$ for the valid ones, and by $[0.7,1.0]$ for the fake ones, as proposed in \cite{salimans2016improved}. 

\noindent\textbf{M2H-GAN.} The generator is identical to the GAN baseline. The discriminator also includes a semi-supervised classification task. Consequently, the output layer is made of $9$ neurons for the $8$ themes of the DECODA framework plus the \textit{FAKE} label. 

Both GAN and M2H-GAN generators are trained to minimize the binary cross-entropy loss observed with the discriminator predictions on their fake generated features, while their discriminators maximize the binary and traditional cross-entropy loss functions of correctly classified sources. Finally, models are trained in an adversarial manner as proposed in \cite{goodfellow2014generative} during $25$ epochs with SGD, no momentum, and with a learning rate set to $0.02$. 

\subsection{Results}
\label{subsec:results}
Two baselines DNN classifiers (Section \ref{subsec:modelsarchitectures}) are tested on both the ASR and TRS versions of the DECODA corpus. Then, GAN-based approaches are trained following Algorithm 1. All the experiments are performed $10$ times and averaged, to alleviate variations due to the random initialization of the parameters.

Table \ref{table:results} reports the average accuracies observed with the GAN, and the more adapted M2H-GAN approaches compared to simple DNN classifiers on the DECODA task. It is first important to note the difference in term of accuracy, between the two baselines during the theme identification on both ASR and TRS transcripts. Indeed, while the standard deviation remains almost equal, both real (\textit{w.r.t} to the validation set) and max test accuracies are different. More precisely, the ASR-based DNN obtains a real test accuracy of $83.4$\% compared to $88.0$\% for the TRS-based DNN, representing a drop of $4.6$\%. This is easily explained by the high WER observed on the ASR transcriptions (Section \ref{subsect:decoda}), that alter significantly the LDA representation and the final classification performances. These results support the initial intuition that a translation of ASR documents to TRS-like representations allow us to better identify the most related theme of a spoken dialogue.

\begin{table}[H]
\centering
\caption{Accuracies obtained by various models on the DECODA corpus. ``Real Test'' stands for the performances observed on the test set w.r.t to the validation set, while ``Max Test'' are the best results obtained. Results are averaged over $10$ runs. The standard deviation is computed over these runs and concern the ``Real Test'' performances.}
\scalebox{0.8}{
    \begin{tabular}{ P{1.7cm}P{0.7cm}P{0.7cm}P{1.3cm}P{1.4cm} P{1.4cm}}
        \hline\hline
        \textbf{Models}&\textbf{Data}& \textbf{Dev.}& \textbf{Real Test}& \textbf{Max Test} & \textbf{Std. Dev.}\\
            \hline
            DNN& TRS &92.5  & 88.0 & 88.5 & 0.016 \\
            DNN& ASR& 89.5  & 83.4 & 84.6 & 0.017\\
            \hline
           	GAN & ASR & 87.0 & 84.1 & 85.2 & 0.012  \\
            \textbf{M2H-GAN} & \textbf{ASR} & \textbf{90.0} & \textbf{85.5} &  \textbf{85.8} & \textbf{0.007} \\
        \hline
    \end{tabular}
    \label{table:results}
}
\end{table}

As a first step to reduce this gap, ASR transcripts inputs are mapped to the TRS ones with a GAN. This approach obtains a best test accuracy of $84.1$\% for ASR inputs, reducing the absolute difference with TRS performances to $3.9$\%. The standard deviation is also lowered to $0.0012$, resulting in a slightly more stable model. Validation performances are altered with an average accuracy of $87.0$\% compared to $89.5$\% and $92.5$\% for the DNNs trained on the ASR and TRS respectively.
     
The Machine-to-Human mapping is then performed with the M2H-GAN. The real test accuracy is increased to $85.5$\%, representing a absolute gain of $1.4$\% and $2.1$\% compared to the simpler GAN and DNN classifier respectively. The gap between the ASR classification performances and the TRS ones is also reduced to $2.5$\%. It is also worth underlying that the standard deviation is halved ($0.007$) in comparison of all the other models, resulting in a more robust representation of the spoken document content. 

\begin{table}[h!]
\centering
\caption{Accuracies obtained by proposed generative models, compared to previous works on the DECODA corpus. ``Real Test'' stands for the performances observed on the test set w.r.t to the validation set, while ``Max Test'' are the best results obtained. Results are averaged over $10$ runs. The standard deviation is computed over these runs and concerns the ``Real Test'' performances }
\scalebox{0.8}{
    \begin{tabular}{ P{1.7cm}P{0.7cm}P{0.7cm}P{1.3cm}P{1.4cm} P{1.4cm}}
        \hline\hline
        \textbf{Models}&\textbf{Data}& \textbf{Dev.}& \textbf{Real Test}& \textbf{Max Test} & \textbf{Std. Dev.}\\
            \hline
            AE\cite{Janod+2016}&ASR& - & 81 & - & -\\
            DAE\cite{Janod+2016}& ASR &-	 &- &74.3 &-\\
            DSAE\cite{Janod+2016}&ASR& 88.0 & 82.0 & 83.0 &-\\
            QDAE\cite{parcollet2017quaternion}&ASR&90.0&85.2&85.2&-\\
            \hline
           	GAN & ASR & 87.0 & 84.1 & 85.2 & 0.012  \\
            \textbf{M2H-GAN} & \textbf{ASR} & \textbf{90.0} & \textbf{85.5} &  \textbf{85.8} & \textbf{0.007} \\
        \hline
    \end{tabular}
    \label{table:results2}
}
\end{table}

Table \ref{table:results2} shows the results observed with GAN and M2H-GAN models compared to previously investigated generative models. Both GAN and M2H outperform the auto-encoders (AE), denoising auto-encoders (DAE), and deep stacked auto-encoders (DSAE) proposed in \cite{Janod+2016,janod2017denoised}. Indeed, encoder and decoder are trained jointly to minimize the reconstruction error, while the generator and discriminator are trained on different objectives impacting on each other. M2H-GAN also give better results than the recent quaternion-valued denoising auto-encoder (QDAE), despite the fact that the QDAE is based on a better document representation and a specific segmentation with the quaternion algebra.
  
\section{Conclusions}

\textbf{Summary.} This paper proposes to use the efficient generative adversarial networks to map an automatically transcribed telephone conversation, to a latent representation of its ``clean'' transcription from human, to better be classified by neural networks. The proposed M2H-GAN, derived from semi-supervised GANs, is compared to common DNN classifiers and a GAN architecture on a realistic task of theme identification of telephone conversations. The M2H-GAN raises the classification accuracy of the noisy ASR transcripts from $83.4$\% for a straightforward DNN to $85.5$\%. The absolute difference with manually transcribed document classification is therefore lowered to $2.5$\%. The model is also more robust with an halved standard deviation over the $10$ runs. 

\noindent\textbf{Future work.} Generative adversarial networks suffer from the fact of being a recent domain area with fewer investigations compared to traditional methods. Therefore, a future work will consist in investigating dedicated GAN models, to better consider the structure of the documents, such as speech turns with recurrent models. Moreover, the instability of GANs training must be investigated in the specific context of noisy document.

\end{document}